\documentclass[10pt, a4paper]{article}

\usepackage[final]{lrec2026} 

\usepackage{multirow}
\usepackage[utf8]{inputenc}
\usepackage[T5]{fontenc} % needed for Vietnamese
\usepackage[vietnamese,english]{babel}
\usepackage{amsmath}
\usepackage{amssymb}  % or amsfonts
\usepackage{microtype}
\usepackage{url}

\UseMicrotypeSet[protrusion]{basicmath} 
\microtypesetup{protrusion=false}

\title{VietJobs: A Vietnamese Job Advertisement Dataset}

\name{Hieu Pham Dinh, Hung Nguyen Huy, Mo El-Haj} 

\address{College of Engineering and Computer Science, VinUniversity \\
     {\small \texttt{\{24hieu.pd, 25hung.nh, elhaj.m\}@vinuni.edu.vn}}\\
}

\abstract{
VietJobs is the first large-scale, publicly available corpus of Vietnamese job advertisements, comprising 48,092 postings and over 15 million words collected from all 34 provinces and municipalities across Vietnam. The dataset provides extensive linguistic and structured information, including job titles, categories, salaries, skills, and employment conditions, covering 16 occupational domains and multiple employment types (full-time, part-time, and internship). Designed to support research in natural language processing and labour market analytics, VietJobs captures substantial linguistic, regional, and socio-economic diversity. We benchmark several generative large language models (LLMs) on two core tasks: job category classification and salary estimation. Instruction-tuned models such as Qwen2.5-7B-Instruct and Llama-SEA-LION-v3-8B-IT demonstrate notable gains under few-shot and fine-tuned settings, while highlighting challenges in multilingual and Vietnamese-specific modelling for structured labour market prediction. VietJobs establishes a new benchmark for Vietnamese NLP and offers a valuable foundation for future research on recruitment language, socio-economic representation, and AI-driven labour market analysis. All code and resources are available at:
\url{https://github.com/VinNLP/VietJobs}.
 \\ \newline }

\begin{document}

\maketitleabstract

% \footnotetext{GitHub Link: \url{https://github.com/dinhieufam/vietjobs}}

\section{Introduction}
Vietnam’s labour market has expanded alongside digital transformation, with online recruitment platforms such as TopCV playing an increasingly central role in connecting job seekers and employers. Recent research highlights how such platforms both reflect and reproduce existing social norms. For instance, explicit references to gender, age, and physical appearance remain common in Vietnamese job postings, influencing wage offers and perceptions of employability \citep{Perroni2023,packard2006gender}. Studies have also shown that job advertisements may privilege particular demographic or aesthetic traits, such as youth or attractiveness, contributing to differentiated opportunities across groups \citep{Perroni2023}. Despite these observations, the linguistic and structural features of Vietnamese recruitment language remain underexplored, in part due to the limited availability of large, well-annotated, and publicly accessible datasets for computational analysis \citep{tran2022practical}.

While previous research on recruitment language has focused mainly on English and other high-resource languages, Vietnamese remains comparatively under-resourced for NLP. The language’s tonal structure, compounding morphology, and frequent code-switching with English present additional challenges for tokenisation, normalisation, and semantic interpretation \citep{Bonoli2012}. This scarcity of data and tools has constrained the development of domain-specific and socially informed NLP applications in Vietnamese. Recent advances in NLP for Vietnamese, including work on job description analysis and fake job detection, have begun to address these challenges, but large, linguistically representative resources remain limited \citep{Vu2025Improving,tran2022practical}.

This paper introduces VietJobs, the first large-scale, open-access dataset of Vietnamese job advertisements. Comprising 48,092 postings collected nationwide, VietJobs integrates linguistic, demographic, and occupational information to provide a structured and diverse view of Vietnam’s online labour market. The dataset supports a wide range of NLP and analytical studies, including linguistic characterisation of recruitment language and benchmarking of state-of-the-art large language models (LLMs) for job category classification and salary estimation under zero-shot, few-shot, and fine-tuned settings \citep{tran2022practical,Vu2025Improving,otani2024natural}. Through these contributions, VietJobs establishes a foundational resource for Vietnamese NLP and labour market analysis. It bridges computational and socio-economic perspectives, offering a platform for studying how language represents professional, educational, and regional variation in recruitment contexts, and supporting future research on language, employment, and AI applications in Southeast Asia \citep{Perroni2023,Bonoli2012,otani2024natural}.

\section{Related Works}

Job advertisements have become a key resource for computational research on labour markets and recruitment language \citep{vogt2023development}. Advances in NLP now enable large-scale analyses of job postings to identify hiring trends, examine linguistic patterns, and estimate salaries or job categories \citep{dawson2020predicting}. However, most existing studies focus on English and other high-resource languages, leaving low- and mid-resource contexts such as Vietnamese largely underrepresented.

Several corpora have supported empirical research on recruitment language. The \textit{Adzuna Global Job Listings} dataset \citep{kanchana_karunarathna_2025} contains over 17,000 English job postings with metadata on compensation and contract type, while the \textit{Djinni Recruitment Dataset} \citep{drushchak-romanyshyn-2024-introducing} includes 150,000 jobs and 230,000 anonymised candidate profiles in English and Ukrainian. Research on recruitment language increasingly explores how wording reflects social and cultural norms in hiring contexts. Studies have examined gendered, age-related, and appearance-based phrasing using approaches ranging from lexicon-based methods to transformer models such as RoBERTa \citep{sharma2025recognizing}, which perform strongly in identifying recurrent linguistic patterns in IT and STEM job markets \citep{KANIJ2024112169}. Work in low-resource languages remains limited, though multilingual initiatives such as the \textit{AraJobs} corpus for Arabic job advertisements \citep{el2025arabjobs} highlight the value of culturally grounded datasets for analysing recruitment discourse.  

VietJobs builds upon this direction by introducing a large-scale, linguistically diverse Vietnamese dataset that enables analysis of language use and representation in Southeast Asian labour markets. Another resource, the \textit{Vietnam Jobs Dataset} available on Kaggle \citep{kaggledataset}, focuses primarily on Vietnamese job titles without including full textual descriptions, which limits its use for NLP-based analysis. Nevertheless, we employ this dataset for comparison purposes, as discussed later in Section~\ref{sec:experiments}.

Salary prediction has also become a key focus of computational labour market research. Prior work has applied regression, ensemble learning, and deep neural models to estimate compensation, showing that variables such as job title, company profile, location, and skills are strong predictors \citep{pluijmaekers2022dataset,bana2022work2vec}. Many datasets include incomplete or inconsistent salary information, limiting model reliability \citep{alsheyab2025jobmarketcheatcodes,el2025arabjobs}. VietJobs mitigates this by including explicit minimum, maximum, and average salary fields in over 70\% of postings, enabling consistent benchmarking for predictive and economic analyses.

Overall, VietJobs extends the linguistic and socio-economic scope of existing Vietnamese resources and complements regionally focused datasets such as \textit{AraJobs}, \textit{SkillSpan}, \textit{JobSkape}, and \textit{Djinni} \citep{el2025arabjobs,zhang2022skillspan,magron-etal-2024-jobskape,drushchak-romanyshyn-2024-introducing}. By capturing the linguistic and cultural specificities of Vietnam, it provides a robust resource for context-sensitive modelling and contributes to the broader advancement of low-resource NLP.

\section{VietJobs Dataset}
\subsection{Data Collection and Corpus Overview}

The \textbf{VietJobs\footnote{\url{https://github.com/VinNLP/VietJobs}}} dataset was compiled from publicly accessible online recruitment platforms in Vietnam during July 2025. Data were collected using the open-source \textit{Crawl4AI} framework \citep{crawl4ai2024}, combined with LLM-assisted parsing through GPT-4o \citep{hurst2024gpt} and Gemini 2.5 \citep{comanici2025gemini25pushingfrontier}, which enabled structured extraction from diverse HTML templates while preserving the linguistic integrity of the text. Access to GPT-4o and Gemini 2.5 was provided through API integration, enabling scalable and efficient parsing and information extraction. Both models were guided using a detailed, task-specific prompt to ensure consistent and schema-compliant outputs. 

All collection procedures were conducted ethically and in compliance with national and institutional regulations, as detailed in Section \ref{sec:ethics}. The crawling process involved two stages: (1) initial URL acquisition, which took approximately 4–5 hours, and (2) full-page crawling and LLM-based information extraction, which required over 7 days. The resulting corpus captures detailed linguistic and structural information from job advertisements representing all 34 provinces and municipalities of Vietnam, with the highest posting volumes in Hanoi and Ho Chi Minh City. A statistical overview of the dataset is presented in Table \ref{tab:dataset_overview}.

\begin{table}[h!]
\centering
\small
\renewcommand{\arraystretch}{1.1}
\resizebox{\columnwidth}{!}{%
\begin{tabular}{l r}
\hline
\textbf{Statistic} & \textbf{Value} \\
\hline
\textbf{General Information} & \\
Total job postings & 48,092 \\
Total tokens (words) & 15,429,581 \\
Average tokens per posting & 321 (mean) \\
Vocabulary size & 78,002 \\
Proportion of English tokens & 0.32\% (49,375 words) \\
Distinct job categories & 16 \\
Collection duration & 1 week \\
\textbf{Geographical Coverage} & \\
Regions covered & 34 provinces \\
Top regions & Hanoi, Ho Chi Minh City \\
\textbf{Salary Information} & \\
Salary range (min–max) & 1–500M VND/month \\
Median (min / avg / max) & 10 / 13 / 15M VND \\
\textbf{Textual Characteristics} & \\
Longest postings & IT \& Digital Engineering (354.8 words) \\
Shortest postings & Languages \& Translation (275.1 words) \\
\textbf{Employment Characteristics} & \\
Contract types & Full-time, Part-time, Internship \\
Benefits / skills fields & Limited coverage \\
\hline
\end{tabular}
}
\caption{Summary statistics of the VietJobs dataset.}
\label{tab:dataset_overview}
\end{table}

With over 15 million words and a vocabulary exceeding 78,000 unique tokens, \textbf{VietJobs} provides an extensive and representative sample of contemporary Vietnamese recruitment discourse. Its geographical and occupational diversity supports both linguistic and computational analyses, enabling the study of bias, code-switching, and socio-economic variation across regions and industries. The unified occupational taxonomy, adapted from ISCO-08, O*NET, and ESCO standards, ensures analytical consistency and facilitates downstream applications such as job classification, salary estimation, and fairness-aware modelling.

\subsection{Occupational Category Distribution}

The \texttt{job\_category} field denotes the occupational domain or functional area represented in each advertisement (e.g., Sales, Engineering, Healthcare). The original categorisation provided by source platforms consisted of 24 distinct labels, many of which exhibited semantic overlap or inconsistent assignment. For example, positions such as \textit{Financial Analyst} and \textit{Accountant} were placed under separate categories, despite belonging to the same broader industrial sector.

To enhance interpretability and comparability, we conducted a systematic category normalisation process. All raw category labels were reviewed and mapped onto a harmonised taxonomy comprising 16 consolidated occupational domains. This approach aligns with established practices in occupational data standardisation in Vietnam \cite{worldbank2019skills}, ensuring analytical consistency while preserving sufficient granularity for downstream NLP tasks. The resulting taxonomy maintains a balance between semantic precision and practical usability, enabling both fine-grained linguistic analyses and macro-level labour market comparisons.

Table~\ref{tab:category_distribution} summarises the distribution of job postings across the 16 categories. The largest segments correspond to \textit{Business, Sales \& Customer Service} (8,276 ads) and \textit{Manufacturing, Manual Labour \& Mechanics} (6,407 ads), reflecting Vietnam’s continued emphasis on commerce, production, and industrial operations. By contrast, specialised fields such as \textit{Agriculture, Energy \& Environment} (322 ads) and \textit{Languages \& Translation} (384 ads) account for smaller shares, indicative of niche professional markets. In total, the dataset comprises 48,092 valid and categorised job advertisements, offering comprehensive coverage of Vietnam’s online recruitment landscape.

\begin{table}[t]
\centering
\small
\renewcommand{\arraystretch}{1.2}
\resizebox{\columnwidth}{!}{%
\begin{tabular}{lc}
\hline
\textbf{Job Category (translated to English)} & \textbf{Ad Count} \\
\hline
Business, Sales \& Customer Service & 8,276 \\
Manufacturing, Manual Labour \& Mechanics & 6,407 \\
Marketing, Communications, Advertising \& Content & 5,950 \\
Finance, Accounting, Banking \& Insurance & 5,530 \\
Tourism, Hospitality \& Services & 4,243 \\
Design, Arts, Entertainment, Media \& Journalism & 3,465 \\
Human Resources, Administration, Legal \& Consulting & 3,276 \\
Construction, Architecture \& Real Estate & 2,811 \\
Information Technology \& Digital Engineering & 1,906 \\
Logistics, Transportation \& Supply Chain & 1,813 \\
Electrical, Electronics \& Telecommunications Engineering & 1,236 \\
Education, Training \& Research & 1,165 \\
Healthcare, Pharmaceuticals \& Biotechnology & 963 \\
Languages \& Translation & 384 \\
Agriculture, Energy \& Environment & 322 \\
Other Occupations & 345 \\
\hline
\textbf{Total} & \textbf{48,092} \\
\hline
\end{tabular}
}
\caption{Distribution of job advertisements across the 16 consolidated occupational categories.}
\label{tab:category_distribution}
\end{table}

\subsection{Salary Distribution}

Salary information in the VietJobs dataset was normalised into three quantitative fields: \texttt{salary\_min}, \texttt{salary\_max}, and \texttt{salary\_avg}, each expressed in millions of Vietnamese Dong (VND) per month. Out of the total 48,092 job postings, 34,365 (71.5\%) explicitly specify a salary range, while the remaining 13,727 (28.5\%) are listed as negotiable. This distribution aligns with prior research on labour market transparency in Southeast Asia, where employers often withhold salary details for competitive or negotiation-related reasons \cite{vanthang2020vietnamese}.

\begin{figure}[h]
    \centering
    \includegraphics[width=\linewidth, height=0.8\linewidth]{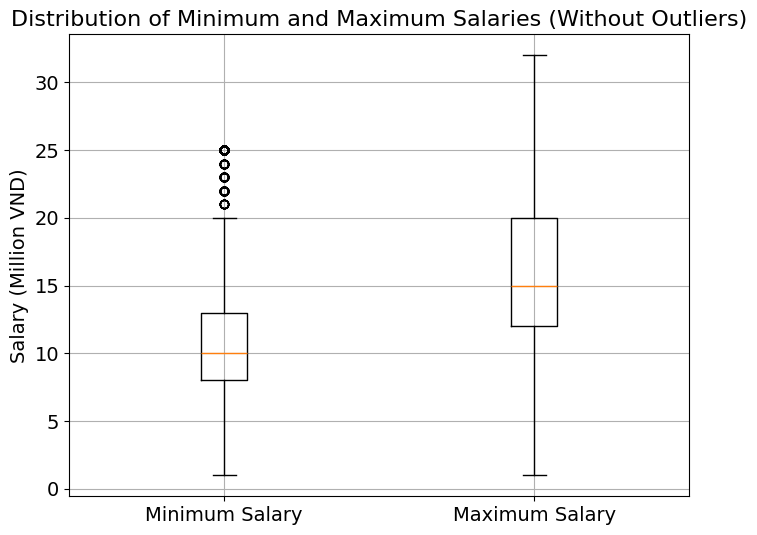}
    \caption{The distribution of Minimum and Maximum Salaries (extreme outliers are removed for better visualisation)}
    \label{fig:salary_dist}
\end{figure}

Figure~\ref{fig:salary_dist} presents the distribution of minimum and maximum salary values after removing extreme outliers. The median minimum salary is approximately 10 million VND, while the median maximum salary is around 15 million VND. This concentration suggests that the majority of advertised positions correspond to entry- or mid-level roles within Vietnam’s labour market. Nonetheless, the upper quartiles extend beyond 30 million VND, indicating the presence of high-paying professional and managerial roles.

The wider spread of maximum salaries relative to minimum salaries highlights substantial wage variability across postings. This variation likely reflects factors such as job seniority, sectoral demand, and skill specialisation. For example, service-oriented occupations—particularly within \textit{Business, Sales \& Customer Service}—tend to exhibit broader salary ranges than technical domains like \textit{Information Technology \& Digital Engineering}. Such disparities suggest that employers frequently adopt flexible salary ranges to appeal to a wider pool of applicants, a strategy commonly observed in dynamic and rapidly expanding economies \cite{lu2023asians}.

The salary distribution exhibits a right-skewed pattern, characterised by a concentration of lower- to mid-income positions alongside a smaller subset of high-income roles. This structure forms the empirical basis for the salary estimation experiments described in Section~\ref{sec:experiments}, where models are tasked with predicting salary values for postings using their key details such as job title or experience required.

\section{Evaluating Job Classification and Salary Estimation with Large Language Models}\label{sec:experiments}

\subsection{Job Category Classification}

The job category classification task focuses on predicting the standardised \texttt{job\_category} label (16 classes) from the textual content of each advertisement, primarily the \texttt{description} field. We evaluate a range of generative large language models (LLMs) available on Hugging Face under three experimental conditions: \textit{zero-shot}, \textit{few-shot}, and \textit{fine-tuned}. All models employ their \texttt{chat} or \texttt{instruct} variants to ensure consistent adherence to natural language instructions and task formatting.

\textbf{Prompting and fine-tuning.}  
In the zero-shot setting, each model receives a concise instruction prompt requesting a single category prediction from the 16 predefined occupational labels. This setup tests a model’s ability to generalise without prior exposure to task-specific examples.
In the few-shot setting, the prompt includes a small number of annotated examples before the test instance, enabling the model to infer task structure and category mappings from limited in-context demonstrations.
For fine-tuning, the models are trained using structured instruction–response pairs, where the prompt contains the job description and the target output corresponds to the correct category label. This conversational-style format allows models to internalise consistent mappings between job descriptions and their respective categories.
During inference, model outputs are restricted to the canonical list of category names. Any responses that deviate from these standard labels are automatically treated as incorrect. This ensures comparability across models and mitigates the influence of generative variability, enabling a fair evaluation of classification accuracy and robustness across instruction-tuned architectures.

\textbf{Evaluation metrics.} Performance is assessed using Accuracy and Macro~F1 scores on the held-out test set. Let $N$ denote the total number of samples, $K$ the number of classes, $\hat{y}_i$ the predicted label, and $y_i$ the ground truth. Accuracy is computed as:
\[
\text{Accuracy} = \frac{1}{N}\sum_{i=1}^{N}\mathbb{I}(\hat{y}_i = y_i)
\]
For each class $k$, Precision, Recall, and F1 are defined as:
\small{\[
\text{Precision}_k = \frac{TP_k}{TP_k + FP_k}
\qquad
\text{Recall}_k = \frac{TP_k}{TP_k + FN_k}
\]
\[
\text{F1}_k = \frac{2 \times \text{Precision}_k \times \text{Recall}_k}{\text{Precision}_k + \text{Recall}_k}
\]
}

The overall Macro~F1 is then given by:
\[
\text{Macro F1} = \frac{1}{K}\sum_{k=1}^{K}\text{F1}_k
\]
Accuracy provides a measure of overall correctness, while Macro~F1 offers a class-balanced evaluation that accounts for performance across both frequent and infrequent job categories.

\subsection{Salary Estimation}

The salary estimation task aims to predict the expected salary range (expressed in the format “X triệu”, which means X million VND) based on structured job information. Each sample includes the following input fields: \texttt{job\_title}, \texttt{contract\_type}, \texttt{location}, \texttt{country}, and \texttt{experience\_required}. We evaluate a set of generative large language models (LLMs) from HuggingFace under three configurations: \textit{zero-shot}, \textit{few-shot} and \textit{fine-tuned}. All models employ their \texttt{chat} or \texttt{instruct} variants to enhance instruction-following capability.

\textbf{Prompting and fine-tuning.} 
In the zero-shot setting, models are prompted to generate a salary prediction in the required “X triệu” format using only the provided job attributes. 
In the few-shot setting, the prompt includes a small number of example pairs of job attributes and their corresponding salary values before the test instance, allowing the model to infer the desired output structure and relationship through in-context learning. 
For the fine-tuned configuration, training instances are structured as instruction–response pairs following the same conversational schema, enabling models to learn the mapping between job attributes and corresponding salary values. 
During inference, the model output is required to strictly adhere to the “X triệu” format. The numeric value $X$ is extracted and compared to the gold-standard salary value. Predictions that are unparsable or deviate from the expected format are considered invalid.

\textbf{Evaluation metrics.} Model performance is evaluated using Root Mean Square Error (RMSE) and the coefficient of determination ($R^2$) on the held-out test set. Let $N$ be the number of samples, $y_i$ the ground-truth salary, and $\hat{y}_i$ the predicted value parsed from the model output. RMSE is computed as:
\[
\text{RMSE} = \sqrt{\frac{1}{N} \sum_{i=1}^{N} (y_i - \hat{y}_i)^2}
\]
and $R^2$ is defined as:
\[
R^2 = 1 - \frac{\sum_{i=1}^{N} (y_i - \hat{y}_i)^2}{\sum_{i=1}^{N} (y_i - \bar{y})^2}
\]
where $\bar{y}$ denotes the mean of the ground-truth salary values. RMSE captures the magnitude of prediction errors, while $R^2$ quantifies the proportion of salary variance explained by the model, providing complementary perspectives on predictive accuracy and generalisation.
\subsection{Model Selection}

To evaluate model performance on Vietnamese job classification and salary estimation, we selected a diverse suite of generative large language models (LLMs) that vary in scale, linguistic coverage, and regional focus. The selection aims to provide a comprehensive comparison across three categories: globally-trained multilingual models, regionally-oriented ASEAN models, and Vietnamese-specialised models.

\textbf{Multilingual models.} We include Qwen2.5-7B-Instruct \cite{qwen25}, Llama-3.1-8B-Instruct \cite{llama31}, Granite-3.3-8B-Instruct \cite{granite33}, and Ministral-8B-Instruct-2410\footnote{\url{https://huggingface.co/mistralai/Ministral-8B-Instruct-2410}}. These models are trained on extensive multilingual corpora and demonstrate robust generalisation across languages. Their ability to process code-mixed or English–Vietnamese inputs makes them suitable for tasks involving international or bilingual job postings commonly observed in Vietnamese recruitment data.

\textbf{ASEAN-focused models.} To assess models tailored for regional linguistic contexts, we evaluate Llama-SEA-LION-v3-8B-IT \cite{llamasealion}, Sailor2-8B-Chat \cite{sailor2}, and SeaLLMs-v3-7B-Chat \cite{seallm3}. These models are explicitly designed to improve coverage of Southeast Asian languages, including Vietnamese, Thai, and Indonesian. Their training objectives and tokenisation strategies better accommodate regional linguistic characteristics, which may enhance performance on language-specific nuances in job advertisements.

\textbf{Vietnamese-specific models.} Finally, we include PhoGPT-4B-Chat \cite{phogpt}, BloomVN-8B-Chat \cite{bloomvn8}, and Vistral-7B-Chat \cite{vistral7}, all of which are trained or fine-tuned predominantly on Vietnamese textual data. These models capture syntactic, lexical, and cultural particularities of Vietnamese, potentially providing superior understanding of contextually rich or idiomatic expressions frequently used in job descriptions.

This configuration enables a systematic comparison between globally multilingual, regionally adapted, and locally specialised models. By evaluating them under both zero-shot and fine-tuned settings, we aim to analyse how linguistic coverage, training focus, and cultural alignment influence LLM performance in Vietnamese recruitment-related tasks.

\subsection{Fine-tuning Configuration}

\textbf{Data Split.} The dataset is partitioned into training, development, and test subsets using an 80\%, 10\%, 10\% ratio. The development set is used for hyperparameter tuning and early stopping, while the final evaluation is conducted exclusively on the held-out test set to ensure fair comparison across models.

\textbf{Fine-tuning Hyperparameters.} Fine-tuning is performed on a single NVIDIA A40 GPU using Low-Rank Adaptation (LoRA) to enable efficient training of large models. The configuration is consistent across all model variants to ensure comparability. The key hyperparameters are as follows:
\begin{itemize}
\item \textbf{LoRA parameters:} rank = 8, $\alpha = 16$, dropout = 0.2
\item \textbf{Target modules:} \texttt{q\_proj}, \texttt{k\_proj}, \texttt{v\_proj}, \texttt{o\_proj}
\item \textbf{Optimiser:} AdamW with learning rate $5\times10^{-5}$
\item \textbf{Batch size:} micro-batch size = 4; effective batch size = 64
\item \textbf{Training epochs:} 2
\item \textbf{Maximum sequence length:} 512 tokens
\item \textbf{Evaluation and checkpoint frequency:} every 200 steps
\item \textbf{Precision:} BF16 enabled
\end{itemize}

\textbf{Compute and Efficiency.} All fine-tuning experiments were executed under identical computational conditions. The job classification fine-tuning process required approximately 5 hours per model, while salary estimation models converged within 1–2 hours. The use of LoRA substantially reduced memory consumption and training time compared to full model fine-tuning, allowing efficient experimentation across multiple model families.

\section{Results and Discussion}

\begin{table*}[t]
\centering
\small
\renewcommand{\arraystretch}{1.2}
\begin{tabular}{lcccc}
\hline
\multirow{2}{*}{\textbf{LLM}} & 
\multicolumn{2}{c}{\textbf{Job Classification}} & 
\multicolumn{2}{c}{\textbf{Salary Estimation}} \\
\cline{2-5}
 & \textbf{Acc} & \textbf{Macro F1} & \textbf{RMSE} & \textbf{R$^{2}$} \\
\hline
\multicolumn{5}{c}{\textbf{Zero-shot}} \\
\hline
Qwen2.5-7B-Instruct & \textbf{\underline{0.31}} & \textbf{\underline{0.32}} & 14.06 & -0.46 \\
Llama-3.1-8B-Instruct & 0.2 & 0.2 & 16.01 & -0.87 \\
Ministral-8B-Instruct-2410 & 0.19 & 0.16 & 40.75 & -13.12 \\
Llama-SEA-LION-v3-8B-IT & 0.26 & 0.29 & \textbf{\underline{11.72}} & \textbf{\underline{0.07}} \\
Sailor2-8B-Chat & 0.16 & 0.16 & 13.74 & -3.29 \\
PhoGPT-4B-Chat & 0 & 0 & - & - \\
% PhoGPT-7B5-Instruct & - & - & - & - \\
Granite-3.3-8B-Instruct & 0 & 0 & 35.37 & -8.19 \\
BloomVN-8B-chat & 0 & 0 & 18.73 & -0.91 \\
SeaLLMs-v3-7B-Chat & 0.09 & 0.07 & 13.14 & -0.28 \\
Vistral-7B-Chat & - & - & 167.07 & -195.89 \\
\hline
\multicolumn{5}{c}{\textbf{Few-shot}} \\
\hline
Qwen2.5-7B-Instruct & \textbf{\underline{0.47}} & \textbf{\underline{0.42}} & 11.45 & 0.03 \\
Llama-3.1-8B-Instruct & 0.42 & 0.36 & 14.73 & -0.60 \\
Ministral-8B-Instruct-2410 & 0.32 & 0.20 & 46.76 & -15.26 \\
Llama-SEA-LION-v3-8B-IT & 0.45 & 0.38 & \textbf{\underline{10.65}} & \textbf{\underline{0.16}} \\
Sailor2-8B-Chat & 0.44 & \textbf{\underline{0.42}} & 11.25 & 0.07 \\
SeaLLMs-v3-7B-Chat & 0.40 & 0.34 & 20.80 & -2.19 \\
\hline
\multicolumn{5}{c}{\textbf{Finetuned}} \\
\hline
Qwen2.5-7B-Instruct & \textbf{\underline{0.34}} & \textbf{\underline{0.33}} & 11.47 & 0.03 \\
Llama-3.1-8B-Instruct & 0.30 & 0.32 & 10.62 & \textbf{\underline{0.17}} \\
Ministral-8B-Instruct-2410 & 0.11 & 0.04 & 12.05 & -0.07 \\
Llama-SEA-LION-v3-8B-IT & 0.30 & \textbf{\underline{0.33}} & \textbf{\underline{10.60}} & \textbf{\underline{0.17}} \\
Sailor2-8B-Chat & 0.30 & 0.30 & 13.31 & -0.30 \\
SeaLLMs-v3-7B-Chat & 0.32 & 0.31 & 10.75 & 0.15 \\
\hline
\end{tabular}
\caption{Performance comparison of various large language models (LLMs) across 2 tasks: Job Classification and Salary Estimation}
\label{tab:results_category_salary}
\end{table*}

\subsection{Job Classification Results}

Table~\ref{tab:results_category_salary} summarises the performance of the evaluated LLMs on the job classification task across three evaluation settings: zero-shot, few-shot, and fine-tuned.

\textbf{Zero-shot performance.} In the absence of task-specific examples, Qwen2.5-7B-Instruct achieves the highest scores, with an accuracy of 0.31 and a Macro~F1 of 0.32. This demonstrates its strong cross-lingual generalisation and instruction-following capability, even without prior exposure to the classification schema. In contrast, models such as PhoGPT-4B-Chat and Granite-3.3-8B-Instruct produce inconsistent or malformed outputs that fail to match the canonical label set, indicating limited understanding of task constraints and label semantics under zero-shot prompting.

\textbf{Few-shot performance.} Incorporating a small number of in-context examples substantially improves performance across all models, with accuracy scores rising to the 0.4x range. Qwen2.5-7B-Instruct again leads with 0.47 accuracy, followed by Llama-SEA-LION-v3-8B-IT (0.45) and Sailor2-8B-Chat (0.44). These results highlight the effectiveness of few-shot prompting in structured text classification, where minimal context allows LLMs to infer task boundaries, label semantics, and linguistic cues from examples. The improvement underscores the adaptability of instruction-tuned models to new domains with limited supervision.

\textbf{Fine-tuned performance.} Parameter-efficient fine-tuning yields moderate gains relative to the zero-shot baseline but does not consistently outperform few-shot prompting. Accuracy values plateau around 0.3x across models, suggesting that while fine-tuning enhances task alignment, it may not fully replicate the contextual reasoning and flexibility afforded by in-context learning. Notably, Ministral-8B-Instruct-2410 exhibits a slight decline in both accuracy and Macro~F1 after fine-tuning, possibly reflecting the overfitting problem of the model.

\textbf{Discussion.} Across all experimental settings, Qwen2.5-7B-Instruct demonstrates the most consistent and robust performance, achieving the best results in both zero-shot and few-shot configurations. The strong results of instruction-tuned multilingual models relative to Vietnamese-specific ones suggest that large-scale multilingual pretraining remains advantageous for generalisable task understanding, even in resource-specific contexts such as Vietnamese recruitment data.

\subsection{Salary Estimation Results}

\begin{table*}[t]
\centering
\small
\renewcommand{\arraystretch}{1.2}
\begin{tabular}{lcccccc}
\hline
\multirow{2}{*}{\textbf{LLM}} & 
\multicolumn{2}{c}{VietJobs} & 
\multicolumn{2}{c}{\textbf{Vietnam Jobs Dataset}} &
\multicolumn{2}{c}{\textbf{Both}} \\
\cline{2-7}
 & \textbf{RMSE} & \textbf{R$^{2}$} & \textbf{RMSE} & \textbf{R$^{2}$} & \textbf{RMSE} & \textbf{R$^{2}$} \\
\hline
\multicolumn{7}{c}{\textbf{Base Model}} \\
\hline
Qwen2.5-7B-Instruct & 14.06 & -0.46 & 16.19 & -0.31  & 15.61 & -0.34 \\
Llama-3.1-8B-Instruct & 16.01 & -0.87 & 1319.94 & -7744.05  & 1092.21 & -6031.91 \\
Ministral-8B-Instruct-2410 & 40.75 & -13.12 & 2600.09 & -30621.95 & 2190.20 & -25135.32 \\
Llama-SEA-LION-v3-8B-IT & \textbf{\underline{11.72}} & \textbf{\underline{0.07}} & \textbf{\underline{15.78}} & \textbf{\underline{-0.05}} & \textbf{\underline{14.61}} & \textbf{\underline{-0.02}} \\
Sailor2-8B-Chat & 13.74 & -3.29 & 99.92 & -12.80  & 85.19 & -12.55 \\
SeaLLMs-v3-7B-Chat & 13.14 & -0.28 & 17.93 & -0.08 & 15.82 & -0.14 \\
\hline
\multicolumn{7}{c}{\textbf{Few-shot}} \\
\hline
Qwen2.5-7B-Instruct & 11.45 & 0.03 & 15.43 & -0.18 & 14.40 & -0.14 \\
Llama-3.1-8B-Instruct & 14.73 & -0.60 & 20.80 & -1.16 & 19.25 & -1.04 \\
Ministral-8B-Instruct-2410 & 46.76 & -15.26 & 45.66 & -9.48 & 45.99 & -10.72 \\
Llama-SEA-LION-v3-8B-IT & \textbf{\underline{10.65}} & \textbf{\underline{0.16}} & \textbf{\underline{14.57}} & \textbf{\underline{-0.06}} & \textbf{\underline{13.40}} & \textbf{\underline{0.01}} \\
Sailor2-8B-Chat & 11.25 & 0.07 & 15.62 & -0.21 & 14.53 & -0.15 \\
SeaLLMs-v3-7B-Chat & 20.80 & -2.19 & 21.78 & -1.37 & 21.50 & -1.54 \\
\hline
\multicolumn{7}{c}{\textbf{Fine-tuned on VietJobs}} \\
\hline
Qwen2.5-7B-Instruct & 11.47 & 0.03 & 14.28 & -0.02  & 13.48 & 0 \\
Llama-3.1-8B-Instruct & 10.62 & \textbf{\underline{0.17}} & 13.93 & 0.03 & 13.08 & 0.06 \\
Ministral-8B-Instruct-2410 & 12.05 & -0.07 & 14.62 & -0.07  & 13.94 & -0.07 \\
Llama-SEA-LION-v3-8B-IT & \textbf{\underline{10.60}} & \textbf{\underline{0.17}} & \textbf{\underline{13.89}} & \textbf{\underline{0.04}}  & \textbf{\underline{13.04}} & \textbf{\underline{0.07}} \\
Sailor2-8B-Chat & 13.31 & -0.30 & 16.84 & -0.41  & 15.97 & -0.40 \\
SeaLLMs-v3-7B-Chat & 10.75 & 0.15 & 14.05 & 0.02  & 13.20 & 0.04 \\
\hline
\multicolumn{7}{c}{\textbf{Fine-tuned on Vietnam Jobs Dataset}} \\
\hline
Qwen2.5-7B-Instruct & 10.77 & 0.15 & 13.40 & 0.10 & 12.70 & 0.11 \\
Llama-3.1-8B-Instruct & 10.82 & 0.14 & 13.34 & 0.11 & 12.68 & 0.12 \\
Ministral-8B-Instruct-2410 & 11.70 & -0.01 & 14.40 & -0.03  & 13.69 & -0.03 \\
Llama-SEA-LION-v3-8B-IT & \textbf{\underline{10.70}} & \textbf{\underline{0.16}} & \textbf{\underline{13.24}} & \textbf{\underline{0.12}} & \textbf{\underline{12.58}} & \textbf{\underline{0.13}} \\
Sailor2-8B-Chat & 10.94 & 0.12 & 13.78 & 0.05 & 13.03 & 0.07 \\
SeaLLMs-v3-7B-Chat & 10.83 & 0.14 & 13.44 & 0.10 & 12.75 & 0.11 \\
\hline
\multicolumn{7}{c}{\textbf{Fine-tuned on Both}} \\
\hline
Qwen2.5-7B-Instruct & 11.32 & 0.06 & 13.45 & 0.10 & 12.88 & 0.09 \\
Llama-3.1-8B-Instruct & 10.31 & 0.22 & 13.20 & \textbf{\underline{0.13}} & 12.44 & 0.15 \\
Ministral-8B-Instruct-2410 & 11.85 & -0.03 & 14.29 & -0.02 & 13.64 & -0.02 \\
Llama-SEA-LION-v3-8B-IT & \textbf{\underline{10.24}} & \textbf{\underline{0.23}} & \textbf{\underline{13.17}} & \textbf{\underline{0.13}} & \textbf{\underline{12.40}} & \textbf{\underline{0.16}} \\
Sailor2-8B-Chat & 12.13 & -0.08 & 13.73 & 0.06 & 13.34 & 0.02 \\
SeaLLMs-v3-7B-Chat & 10.48 & 0.19 & 13.41 & 0.10 & 12.64 & 0.12 \\
\hline
\end{tabular}
\caption{Performance comparison of LLMs in Salary Estimation with different settings}
\label{tab:results_salary_both}
\end{table*}

Table~\ref{tab:results_salary_both} summarises the performance of the evaluated LLMs on the salary estimation task under five experimental settings: zero-shot, few-shot, fine-tuned on VietJobs, fine-tuned on the Vietnam Jobs Dataset \cite{kaggledataset}, and fine-tuned on the combined datasets. Overall, model performance improves consistently as task-specific supervision is introduced through fine-tuning. For example, \textbf{Qwen2.5-7B-Instruct} achieves an RMSE of 14.06 and an $R^2$ of $-0.46$ in the zero-shot setting on VietJobs, which improves to an RMSE of 11.32 and an $R^2$ of 0.06 after fine-tuning on both datasets. Comparable trends are observed across the other models, highlighting the importance of domain adaptation in salary prediction.

Across all configurations, \textbf{Llama-SEA-LION-v3-8B-IT} delivers the strongest and most consistent results. Even without explicit task conditioning, it achieves an RMSE of 11.72 and an $R^2$ of 0.07 in the zero-shot scenario on VietJobs, outperforming all other models. With few-shot prompting, its performance further improves to 10.65 RMSE and 0.16 $R^2$. After fine-tuning on both datasets, the model maintains robust performance with an RMSE of 12.40 and $R^2$ of 0.16 when evaluated on the combined data. This pattern suggests that Llama-SEA-LION effectively integrates pre-trained multilingual knowledge with task-specific cues, allowing it to generalise well across both datasets.

The observed performance gains from fine-tuning on the Vietnam Jobs Dataset and the combined data can be attributed to increased data diversity and representativeness. Compared with VietJobs, the Vietnam Jobs Dataset provides a broader range of industries, salary brackets, and contextual descriptors, offering richer information for modelling salary patterns. Fine-tuning on the combined corpus exposes models to this wider distribution of job contexts, leading to lower RMSE values and higher $R^2$ scores—indicative of improved predictive accuracy and generalisability.

In summary, the results reveal three key findings: (1) task-specific fine-tuning substantially enhances model accuracy for salary estimation; (2) performance generally follows the trend \textit{zero-shot < few-shot < fine-tuned on VietJobs < fine-tuned on Vietnam Jobs Dataset < fine-tuned on both datasets}; and (3) \textbf{Llama-SEA-LION-v3-8B-IT} emerges as the most robust and effective model for this task, demonstrating strong adaptability and generalisation across different data sources.

\section{Limitations}

While VietJobs represents the largest publicly available dataset of Vietnamese job advertisements, several limitations remain. It is sourced from a single online recruitment platform called TopCV, which may not reflect all sectors or informal employment in Vietnam, resulting in underrepresentation of some industries or job types. The postings also follow the platform’s linguistic and structural conventions, which could introduce systematic biases.

Salary information, though common, is not consistently standardised and may involve rounding or omission. Textual fields such as job descriptions sometimes contain duplicated or templated content, potentially affecting linguistic analyses and model outcomes. This study focuses on two core tasks—job category classification and salary estimation—using general-purpose large language models. These provide initial benchmarks but do not capture the full scope of possible NLP applications. Finally, although the dataset excludes identifiable personal data, downstream users should still apply appropriate ethical and legal safeguards.

\section{Conclusion and Future Work}

This paper introduced VietJobs, a large-scale dataset of 48,092 Vietnamese job advertisements for research in job classification, salary estimation, and labour market analysis. The dataset covers 16 normalised job categories and includes structured fields such as job titles, salaries, skills, and employment conditions. Its utility was evaluated through two core tasks using large language models, with instruction-tuned systems such as Qwen2.5-7B-Instruct and Llama-SEA-LION-v3-8B-IT achieving the most consistent results. Fine-tuning on combined datasets produced the most accurate salary estimates.

VietJobs provides a foundation for future research in computational labour market monitoring and recruitment modelling. Future work may expand coverage across additional platforms and time periods, incorporate multilingual or demographic data, and explore advanced modelling methods such as retrieval-augmented generation or domain-adaptive pretraining. In addition, while this study focused exclusively on the performance of large language models, future research should conduct systematic comparisons with traditional machine learning approaches, such as TF-IDF feature representations combined with classifiers like Logistic Regression. Such comparisons would help quantify the relative benefits of LLM-based methods over conventional baselines in Vietnamese job classification and salary prediction tasks.

\section*{Ethical Considerations and Limitations}\label{sec:ethics}
All data used in this study were collected from publicly accessible pages of the \textit{TopCV.vn} website, which does not prohibit web scraping under its \texttt{robots.txt} policy. Only non-personal, publicly available job advertisement content was included; no user profiles, résumés, or private materials were accessed. The dataset contains no personally identifiable information (PII), and all text was processed solely for research purposes in accordance with fair-use and data protection principles.
This study received ethical approval from an Institutional Ethical Review Board under the principal investigator: Mo El-Haj. The approval (Decision No.~J/2025/CN/HDDD) was granted in accordance with Circular 43/2024/TT-BYT, which regulates the establishment and operation of ethics councils in research, alongside relevant institutional policies. The study was classified as minimal risk and authorised for the period October–December 2025. All research activities adhered to institutional guidelines, Vietnamese national regulations, and internationally recognised ethical standards, including the principles of the Declaration of Helsinki. No personal or sensitive data were collected, and no human subjects were involved.

\bibliographystyle{lrec2026-natbib}
\bibliography{lrec2026}

% \section{Language Resource References}\label{lr:ref}

% \bibliographystylelanguageresource{lrec2026-natbib}
% \bibliographylanguageresource{languageresource}

\end{document}